\pdfoutput=1
\documentclass[final]{hld2026}

\usepackage{microtype}
\usepackage{booktabs}
\usepackage{mathtools}
\usepackage{algorithm}
\usepackage{algorithmic}
\usepackage{xcolor}
\usepackage[capitalize,noabbrev]{cleveref}

\makeatletter
\renewcommand*{\@titlefoot}{}
\makeatother

\crefname{proposition}{Proposition}{Propositions}
\Crefname{proposition}{Proposition}{Propositions}
\crefname{algorithm}{Algorithm}{Algorithms}
\Crefname{algorithm}{Algorithm}{Algorithms}

\newcommand{\R}{\mathbb{R}}
\newcommand{\E}{\mathbb{E}}
\newcommand{\bx}{\mathbf{x}}
\newcommand{\ba}{\mathbf{a}}
\newcommand{\bw}{\mathbf{w}}
\newcommand{\bF}{\mathbf{F}}
\newcommand{\bP}{\mathbf{P}}
\newcommand{\camera}[1]{#1}
\DeclareMathOperator{\diag}{diag}
\DeclareMathOperator{\sign}{sign}

\title[Optimizer Dependence of Neural Scaling Laws]{On the Optimizer Dependence of Neural Scaling Laws}

\hldauthor{\Name{Vansh Ramani\nametag{*}}\\
\addr{Department of Computer Science and Engineering, Indian Institute of Technology Delhi}\\
\AND
\Name{Shourya Vir Jain\nametag{*}}\\
\addr{Department of Electrical Engineering, Indian Institute of Technology Delhi}\\[0.5ex]
\addr{\normalfont *Equal contribution}}

\begin{document}

\maketitle

\begin{center}
{\small Accepted at the 4th Workshop on High-dimensional Learning Dynamics (HiLD), ICML 2026.}
\end{center}

%% ====================================================================
%% ABSTRACT
%% ====================================================================
\begin{abstract}
The scaling exponent $\alpha$ in neural scaling laws $L(N) \propto N^{-\alpha}$ is commonly treated as a fixed constant set by architecture and data.
We present evidence that $\alpha$ depends systematically on the optimizer.
In controlled random-feature regression experiments---the canonical theoretical framework for neural scaling---we measure $\alpha$ across five optimizer variants and six spectral conditions.
Preconditioned optimizers consistently yield steeper scaling (larger $\alpha$), \camera{with the $\alpha$-shift increasing across most of the tested spectral range, peaking near $s=1.5$, and remaining large at $s=2.0$.}
At $s \approx 1.0$ (characteristic of natural language), the full natural gradient achieves $\alpha \approx 0.31$ versus $\alpha \approx 0.12$ for gradient descent---\camera{a $2.6\times$ larger fitted exponent} that, within the random-feature model, compounds with each model-size doubling.
Whether and how this exponent shift transfers to large-scale LLM training---where recent evidence suggests the advantage may attenuate with scale---remains an important open question.
Our results imply that scaling-law forecasts should account for optimizer choice, and we provide a spectral diagnostic predicting when advanced optimizers will pay off.
\end{abstract}

%% ====================================================================
%% 1. INTRODUCTION
%% ====================================================================
\section{Introduction}
\label{sec:intro}

Neural scaling laws---the observation that test loss decreases as a power law $L(N) \propto N^{-\alpha}$ with model size $N$---govern how frontier laboratories allocate compute \citep{kaplan2020scaling, hoffmann2022training}.
The exponent $\alpha$ determines returns to scaling: larger $\alpha$ means each doubling of parameters yields greater loss reduction.
Current practice treats $\alpha$ as a fixed constant fit from a few training runs, then extrapolated to set budgets for orders-of-magnitude larger models.

Yet recent results suggest $\alpha$ may not be fixed.
\citet{chen2025second} showed that second-order optimizers---Muon, SOAP \citep{vyas2024soap}, and Shampoo \citep{gupta2018shampoo}---can deliver $\sim\!1.4\times$ compute speedups over AdamW when their hyperparameters are scaled carefully.
\citet{liu2025muon} confirmed this at frontier scale: Muon matches AdamW at 52\% of training FLOPs.
However, the picture is not uniformly positive: \citet{wen2025fantastic} report that the Muon/SOAP speedup over AdamW \emph{decays} with model size, dropping to ${\sim}1.1\times$ at 1.2B parameters---suggesting the effect may attenuate at scale.
These conflicting findings motivate a more precise question: does Muon improve the \emph{constant factor} in the scaling law, or does it change the \emph{exponent} itself?  And if the exponent shifts, does the shift persist, shrink, or vanish at scale?

This distinction matters enormously.
A constant-factor improvement saves a fixed proportion of compute.
An exponent improvement, if it persists, \emph{compounds}: the advantage grows with every doubling.
If preconditioned optimizers change $\alpha$---even partially---then forecasts calibrated under AdamW may underestimate returns to scaling under Muon.

We investigate this question in the random-feature regression framework of \citet{maloney2022solvable} and \citet{bordelon2024dynamical}, where data spectrum, optimizer preconditioner, and model capacity can be cleanly isolated.
This framework---rooted in the natural gradient theory of \citet{amari1998natural} and its modern extensions \citep{martens2020new}---lets us study how preconditioning interacts with spectral structure to determine scaling exponents.

\textbf{Contributions.}
We make four contributions:
\textbf{(1)} the first systematic measurement of $\alpha$ as a joint function of optimizer type and data spectral decay, across five preconditioner variants and six spectral conditions (\cref{sec:results});
\textbf{(2)} a mechanistic explanation via an informal proposition: preconditioned optimizers equalize per-mode learning rates, increasing well-learned modes at given $N$ (\cref{sec:setup});
\textbf{(3)} \camera{a comparison to published optimizer-scaling results, treated as qualitative external context rather than a parameter-by-parameter fit} (\cref{sec:validation}); and
\textbf{(4)} a falsifiable prediction: the $\alpha$-shift \camera{should generally grow with spectral decay before finite-size or finite-budget saturation}; on flat-spectrum data ($s < 0.3$), preconditioned optimizers should provide negligible scaling improvement (\cref{sec:discussion}).

%% ====================================================================
%% 2. SETUP
%% ====================================================================
\section{Setup: Random-Feature Regression with Optimizer Preconditioning}
\label{sec:setup}

\subsection{The Random-Feature Scaling Model}

We adopt the random-feature regression framework of \citet{maloney2022solvable}, which provides a solvable model for neural scaling laws.

\textbf{Data distribution.}
Inputs $\bx \in \R^D$ are drawn from $\mathcal{N}(0, \Sigma)$ with eigenvalues $\lambda_i = i^{-(1+s)}$, $i = 1, \ldots, D$, where $s > 0$ is the \emph{spectral exponent}.
Natural language embeddings typically exhibit $s \approx 0.8\text{--}1.2$ based on Zipfian frequency distributions \citep{kaplan2020scaling, bahri2024explaining}.

\textbf{Teacher function.}
$f^*(\bx) = \sum_{k=1}^{K^*} v_k \, \sigma(\bw^*_k \cdot \bx)$, with ReLU activation $\sigma$ and teacher coefficients $v_k^2 \propto k^{-b}$ (source exponent $b$).

\textbf{Student model.}
$\hat{y}(\bx) = \sum_{j=1}^N a_j \, \sigma(\bw_j \cdot \bx)$, where $\{\bw_j\}$ are fixed random weights from $\mathcal{N}(0, I_D/D)$ and trainable parameters are $\ba \in \R^N$.
The student minimizes $L = \E[(\hat{y}(\bx) - f^*(\bx))^2]$.

This produces power-law scaling $L(N) \propto N^{-\alpha}$ where $\alpha$ depends on $s$ and $b$ \citep{maloney2022solvable}.
Crucially, the top-layer problem is a linear regression, allowing different optimizers to be implemented as \emph{preconditioners} with no confounding from feature learning.

\subsection{Optimizer Variants as Preconditioners}

We implement five variants, each preconditioning the gradient $\nabla_{\ba} L$:
\begin{itemize}
\setlength{\itemsep}{0pt}
\setlength{\parskip}{0pt}
    \item \textbf{GD} (baseline): $\bP = I$.
    \item \textbf{Diagonal} (AdamW proxy): $\bP = \diag(\bF^\top \bF)^{-1/2}$.
    \item \textbf{Full NG} (Shampoo/K-FAC proxy): $\bP = (\bF^\top \bF)^{-1}$; equivalent to solving the normal equations \citep{amari1998natural, martens2020new}.
    \item \textbf{Sign-GD}: $\ba \leftarrow \ba - \eta \, \sign(\nabla L)$.
    \item \textbf{Matrix-Sign} (Muon proxy): $\bP = (\bF^\top \bF)^{-1/2}$; equalizes singular values, analogous to Muon's Newton-Schulz orthogonalization \citep{jordan2024muon}.
\end{itemize}
Here $\bF \in \R^{n \times N}$ is the random-feature matrix on training data.

\subsection{Spectral Intuition: Why Preconditioning Shifts \texorpdfstring{$\alpha$}{alpha}}

Under GD with power-law data, learning proceeds mode-by-mode from highest variance downward.
At model size $N$, GD effectively resolves $O(N^{\gamma})$ spectral modes for $\gamma < 1$ set by $s$ and $b$.
A preconditioner equalizes convergence rates, rescuing ``underserved'' low-variance modes.
We state this as a spectral argument (not a formal theorem; a fully rigorous proof would require resolvent-trace analysis in the style of \citealt{bordelon2024dynamical}):

\begin{proposition}[Spectral Heuristic---Informal]
\label{prop:spectral}
Consider random-feature regression with covariance $\lambda_i \propto i^{-(1+s)}$ and preconditioner $\bP$.
Let $\alpha_{\textup{GD}}(s)$ and $\alpha_{\bP}(s)$ denote the scaling exponents under GD and preconditioned GD respectively.
\camera{Heuristically:} \textup{(i)} $\alpha_{\bP}(s) \geq \alpha_{\textup{GD}}(s)$ for any positive-definite $\bP$ that downweights high-variance modes relative to low-variance modes; \textup{(ii)} the gap $\Delta\alpha(s) = \alpha_{\bP}(s) - \alpha_{\textup{GD}}(s)$ \camera{is expected to grow with $s$ until finite-size or finite-budget effects become important}; and \textup{(iii)} $\Delta\alpha(s) \to 0$ as $s \to 0$ (flat spectrum).
\end{proposition}

\emph{Supporting argument.} When $\bP$ fully inverts the covariance (full NG), every mode converges at the same rate, so all $N$ capacity units are used uniformly.
Under GD, the effective capacity per mode decays with the eigenvalue, so the bottom modes remain under-learned.
The fraction of ``wasted'' capacity grows with spectral steepness $s$, making the $\alpha$-gap larger.
When the spectrum is flat ($s \to 0$), GD already treats modes equally, so preconditioning adds nothing.
\Cref{app:prop-discussion} provides a detailed per-mode convergence-rate derivation supporting all three claims, including explicit convergence factors under GD, full NG, and Matrix-Sign (\cref{eq:gd-convergence,eq:ng-convergence}).

%% ====================================================================
%% 3. SIMULATION RESULTS
%% ====================================================================
\section{Simulation Results}
\label{sec:results}

\subsection{Experimental Protocol}

We fix dimension $D\!=\!1000$, teacher complexity $K^*\!=\!100$, source exponent $b\!=\!1.0$, and sweep six spectral conditions ($s$ from $0.25$ to $2.0$) and eight model sizes ($N$ from $25$ to $5{,}000$) across all five optimizers.
Each configuration uses 10~random seeds, $T\!=\!2000$ gradient steps, and \camera{optimizer-specific step-size rules} (\cref{app:details}).
Training set size scales with $N$; test set size is $5000$.
Scaling exponents are fit by OLS on $(\log N, \log L)$ for $N \geq 200$.

\subsection{Core Results}

\begin{figure}[t]
    \centering
    \includegraphics[width=0.78\textwidth]{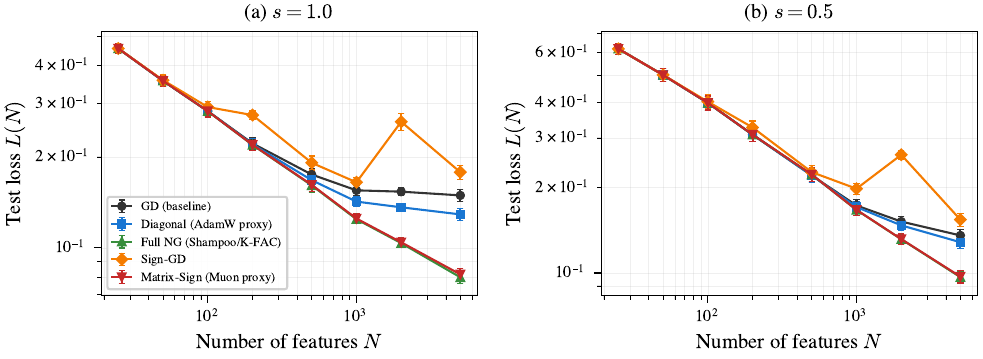}
    \caption{Scaling curves $L(N)$ on log-log axes for five optimizer variants.
    \textbf{(a)}~At spectral exponent $s = 1.0$ (steep spectrum, characteristic of natural language), curves fan out: preconditioned optimizers achieve lower loss at every $N$ with visibly steeper slopes.
    \textbf{(b)}~At $s = 0.5$ (flatter spectrum), the fan-out narrows, confirming that the optimizer advantage depends on spectral steepness.
    Shaded bands show $\pm 1$ SE over 10 random seeds per configuration.}
    \label{fig:scaling-curves}
    \vskip -0.1in
\end{figure}

\Cref{fig:scaling-curves} shows the core phenomenon.
At $s = 1.0$ (panel a), the full natural gradient achieves substantially lower loss at every $N$ with visibly steeper slope.
At $s = 0.5$ (panel b), optimizers produce nearly parallel curves.

\begin{figure}[t]
    \centering
    \includegraphics[width=0.62\textwidth]{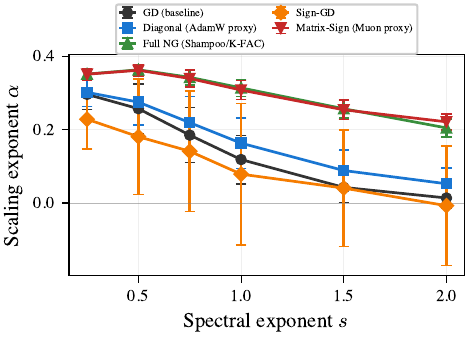}
    \vskip -0.05in
    \caption{Scaling exponent $\alpha$ vs.\ spectral exponent $s$ for all five optimizers.
    Full NG and Matrix-Sign maintain $\alpha > 0.2$ even at $s = 2.0$, while GD collapses to $\alpha \approx 0.01$.
    \camera{The large growth of the $\alpha$-gap over most of the range, followed by mild saturation at the steepest spectra, supports Proposition~\ref{prop:spectral}.}
    Error bars: 95\% CIs from OLS regression on $(\log N, \log L)$.}
    \label{fig:alpha-vs-s}
    \vskip -0.15in
\end{figure}

\Cref{fig:alpha-vs-s} quantifies the main result. Key patterns:

\textbf{Preconditioning improves scaling.}
At every $s$, $\alpha_{\text{FullNG}} > \alpha_{\text{GD}}$, with consistent ordering $\alpha_{\text{FullNG}} \approx \alpha_{\text{MatSign}} > \alpha_{\text{Diag}} > \alpha_{\text{Sign}} \approx \alpha_{\text{GD}}$.
Full NG and Matrix-Sign achieve nearly identical $\alpha$ (e.g., $0.314$ vs.\ $0.308$ at $s = 1.0$), consistent with both being strong spectral preconditioners.

\textbf{The $\alpha$-shift grows with spectral steepness.}
At $s = 0.25$, $\Delta\alpha = 0.055$; at $s = 1.0$, $\Delta\alpha = 0.195$; \camera{at $s = 1.5$, $\Delta\alpha = 0.214$; and at $s = 2.0$, $\Delta\alpha = 0.191$.}
GD's exponent \emph{collapses} as $s$ grows ($\alpha_{\text{GD}}$: $0.30 \to 0.01$), while preconditioned optimizers degrade gracefully ($\alpha_{\text{FullNG}}$: $0.35 \to 0.20$).
\camera{This broadly supports Proposition~\ref{prop:spectral}, with saturation at the steepest tested spectra.}

\textbf{\camera{Effective multiplier is a qualitative comparison.}}
\Cref{fig:compute-mult} translates the $\alpha$-shift into effective compute multipliers.
\camera{At $s \approx 1.0$, the random-feature model predicts a substantial parameter-equivalent advantage for strong preconditioners; because this quantity is not a wall-clock or FLOP-normalized speedup, we use published Muon results as directional context rather than a quantitative target.}

\Cref{tab:alpha} reports full numerical results.
Power-law fit quality is high: $R^2 > 0.95$ for all (optimizer, $s$) pairs except Sign-GD at large $s$, where the high-variance updates produce noisy scaling (\cref{app:r2}).

\subsection{Interpretation}

The scaling exponent $\alpha$ measures how efficiently additional capacity reduces loss.
Under GD with power-law data, capacity is allocated inefficiently: the optimizer ``overspends'' on high-variance modes and ``underspends'' on low-variance modes that still contain signal.
A preconditioner corrects this by equalizing effective learning rates.
This is not merely a convergence-speed effect---the preconditioner changes the \emph{asymptotic} relationship between $N$ and $L$ because it changes which modes contribute to residual loss at each $N$.

%% ====================================================================
%% 4. VALIDATION
%% ====================================================================
\section{Validation Against Published Results}
\label{sec:validation}

The random-feature model omits feature learning, depth, attention, and tokenization, so we treat real-model results as a qualitative check rather than a parameter-by-parameter fit.
\citet{chen2025second} studied transformer models from roughly 190M to 1.4B parameters and showed that carefully scaled second-order optimizers, including Muon, SOAP, and Shampoo, can retain about a ${\sim}1.4\times$ compute advantage over AdamW.
\camera{That reported Muon advantage is directionally aligned with our finding that, at $s \approx 1.0$, Matrix-Sign achieves a large positive shift ($\alpha_{\text{MatSign}}-\alpha_{\text{GD}}\approx0.19$).}
\camera{We do not interpret the random-feature multiplier as a calibrated prediction of transformer FLOP savings; the important agreement is directional: spectral preconditioning helps most in steep-spectrum regimes.}

Our findings also connect to recent theory.
Prior random-feature and statistical-mechanics analyses show that $\alpha$ is governed by the data spectrum and target-function structure \citep{maloney2022solvable,bordelon2024dynamical,bahri2024explaining}; we show that the optimizer also changes the effective spectrum.
\citet{bahri2024explaining} derive scaling exponents from statistical mechanics with a fixed optimizer, while \citet{bergsma2025power} show regularization and data composition can shift $\alpha$.
Our contribution is complementary: the optimizer preconditioner itself reshapes the effective spectrum.
A key caveat is the scale-dependent decay reported by \citet{wen2025fantastic}, where Muon/SOAP gains over AdamW shrink from ${\sim}1.4\times$ at small scales to ${\sim}1.1\times$ at 1.2B parameters.
This suggests the exponent shift may attenuate with feature learning or finite-$N$ effects; \cref{app:scale-decay} discusses these mechanisms.

%% ====================================================================
%% 5. DISCUSSION
%% ====================================================================
\section{Discussion and Limitations}
\label{sec:discussion}

\textbf{Practical diagnostic.}
Our results yield a simple rule: (1)~measure your data's spectral decay from the embedding-covariance eigenvalues; (2)~if $s$ is large, preconditioned optimizers improve $\alpha$, not just convergence speed; (3)~if $s$ is small, optimizer choice matters less for scaling.

\textbf{Falsifiable prediction.}
Training language models with Muon and AdamW on an artificially flattened-spectrum dataset (e.g., domain-equalized or whitened embeddings) should cause the Muon advantage to shrink or vanish.

\textbf{Limitations.}
(1)~The random-feature model is lazy and omits feature learning; relative optimizer comparisons should be more robust than absolute $\alpha$ values.
(2)~Source exponent $b$ is fixed at $1.0$ in the main experiments (see \cref{app:robustness} for $b=2.0$).
(3)~Our Muon proxy captures spectral equalization but omits momentum and normalization, and we study width scaling only.
(4)~The evidence from \citet{wen2025fantastic} suggests the shift may attenuate at scale, which our current framework does not yet explain.

\textbf{Proposed real-model validation.}
The direct test is to train GPT-2--scale transformers with AdamW and Muon on standard web text and a spectrum-flattened variant, then fit $\alpha_{\text{Muon}}-\alpha_{\text{AdamW}}$ across model sizes.
This exceeds the present workshop budget; \cref{app:proposed-experiment} gives a concrete protocol.

\textbf{Future work.}
Priorities are a closed-form derivation of $\alpha(s,b,\bP)$, a two-layer feature-learning extension, and the real-model validation above.
The framework may also connect to Edge-of-Stability dynamics \citep{cohen2021gradient, kalra2026scalable}.

\noindent\textbf{LLM use disclosure.} LLMs were used only for paragraph-level reframing and literature review support.

\begingroup
\footnotesize
\setlength{\bibsep}{1pt}
\bibliography{references}
\endgroup

%% ====================================================================
%% APPENDIX
%% ====================================================================
\newpage
\appendix
\onecolumn

{\small
\noindent\textbf{Appendix roadmap.}
\cref{app:details}~details \textbar{}
\cref{app:full-results}~full results \textbar{}
\cref{app:algorithm}~algorithm \textbar{}
\cref{app:prop-discussion}~spectral derivation \textbar{}
\cref{app:r2}~fit quality \textbar{}
\cref{app:robustness}~robustness \textbar{}
\cref{app:convergence}~convergence \textbar{}
\cref{app:sign}~Sign-GD \textbar{}
\cref{app:caveats}~caveats \textbar{}
\cref{app:scale-decay}~scale decay \textbar{}
\cref{app:proposed-experiment}~validation protocol \textbar{}
\cref{app:code}~code.
\par}

\section{Full Numerical Results}
\label{app:full-results}

% Auto-generated Table 1 -- resized for the official one-column HiLD style.
\begin{table}[!htbp]
  \caption{Scaling exponents $\alpha$ ($\pm$ 95\% CI) across optimizer variants and spectral exponents $s$, fit by OLS on $(\log N, \log L)$ for $N \geq 200$ over 10 seeds.
  Larger $\alpha$ indicates steeper (more favorable) scaling.
  \textbf{Bold}: best $\alpha$ per column.
  Full NG and Matrix-Sign dominate at every $s$; \camera{the gap over GD generally widens with $s$ before mild saturation at the steepest spectra.}}
  \label{tab:alpha}
  \centering
  \begin{scriptsize}
  \setlength{\tabcolsep}{4pt}
  \renewcommand{\arraystretch}{1.1}
  \resizebox{\textwidth}{!}{%
  \begin{tabular}{lcccccc}
    \toprule
    Optimizer & $s=0.25$ & $s=0.5$ & $s=0.75$ & $s=1.0$ & $s=1.5$ & $s=2.0$ \\
    \midrule
    GD (baseline) & $0.296 \pm 0.041$ & $0.257 \pm 0.068$ & $0.185 \pm 0.075$ & $0.118 \pm 0.066$ & $0.043 \pm 0.041$ & $0.014 \pm 0.027$ \\
    Diagonal (AdamW proxy) & $0.302 \pm 0.039$ & $0.275 \pm 0.062$ & $0.219 \pm 0.071$ & $0.163 \pm 0.069$ & $0.089 \pm 0.056$ & $0.052 \pm 0.043$ \\
    Full NG (Shampoo/K-FAC) & $\mathbf{0.351} \pm 0.015$ & $\mathbf{0.363} \pm 0.015$ & $\mathbf{0.343} \pm 0.021$ & $\mathbf{0.314} \pm 0.023$ & $\mathbf{0.257} \pm 0.026$ & $0.205 \pm 0.025$ \\
    Sign-GD & $0.228 \pm 0.081$ & $0.181 \pm 0.158$ & $0.141 \pm 0.164$ & $0.079 \pm 0.192$ & $0.041 \pm 0.159$ & $-0.007 \pm 0.163$ \\
    Matrix-Sign (Muon proxy) & $\mathbf{0.351} \pm 0.015$ & $0.362 \pm 0.015$ & $0.339 \pm 0.022$ & $0.308 \pm 0.026$ & $0.254 \pm 0.026$ & $\mathbf{0.221} \pm 0.022$ \\
    \bottomrule
  \end{tabular}%
  }
  \end{scriptsize}
  \vskip -0.1in
\end{table}

\begin{figure}[t]
    \centering
    \includegraphics[width=0.62\textwidth]{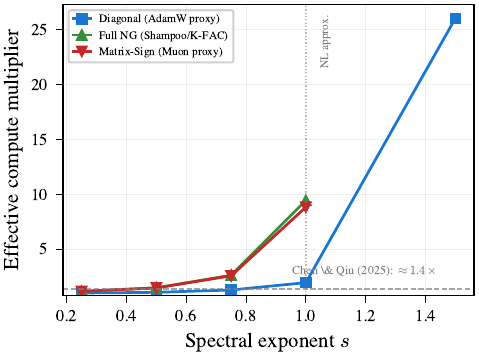}
    \vskip -0.05in
    \caption{Effective compute multiplier (how many $\times$ more GD parameters are needed to match preconditioned loss) vs.\ $s$.
    \camera{Dashed horizontal line: the ${\sim}1.4\times$ Muon speedup reported by \citet{chen2025second}, shown only as a qualitative reference because the plotted quantity is parameter-equivalent rather than FLOP-equivalent.}
    Vertical dotted line: approximate spectral exponent of natural language data ($s \approx 1.0$).
    \camera{The multiplier grows rapidly with $s$ for strong preconditioners and should not be read as a direct wall-clock forecast.}}
    \label{fig:compute-mult}
    \vskip -0.15in
\end{figure}

\section{Experimental Details}
\label{app:details}

\subsection{Hyperparameters}
\label{app:hyperparams}

\Cref{tab:hyperparams} summarizes all optimizer hyperparameters.
For iterative optimizers, \camera{GD, Diagonal, and Matrix-Sign use deterministic spectral step-size rules, while Sign-GD uses a small candidate search.}
Weight decay is zero throughout.
Training runs for $T = 2000$ gradient steps; Full~NG uses a direct solve (one-step Newton via Cholesky decomposition with Tikhonov regularization $\lambda_{\text{reg}} = 10^{-6}$), so the step count is irrelevant for that optimizer.

\begin{table}[ht]
  \caption{Optimizer hyperparameters.
  $\lambda_{\max}$ denotes the largest eigenvalue of $\bF^\top\bF / n$.
  \camera{The iterative step sizes are those used in the released simulation script; no held-out validation split is used for GD, Diagonal, or Matrix-Sign.}}
  \label{tab:hyperparams}
  \centering
  \begin{small}
  \setlength{\tabcolsep}{5pt}
  \renewcommand{\arraystretch}{1.1}
  \begin{tabular}{llll}
    \toprule
    Optimizer & LR grid & Selection criterion & Other \\
    \midrule
    GD & \camera{$1.5/\lambda_{\max}$} & \camera{Fixed spectral rule} & --- \\
    Diagonal & \camera{$1.5/\lambda_{\max}(\diag(\bP)\bF^\top\bF/n)$} & \camera{Fixed spectral rule} & $\bP = \diag(\bF^\top\bF)^{-1/2}$ \\
    Full NG & N/A (direct solve) & --- & Tikhonov $\lambda_{\text{reg}} = 10^{-6}$ \\
    Sign-GD & \camera{$\{10^{-4}, 10^{-3}, 10^{-2}\}$} & Lowest residual norm & --- \\
    Matrix-Sign & \camera{$1.5/\sqrt{\lambda_{\max}}$} & \camera{Fixed spectral rule} & $\bP = (\bF^\top\bF)^{-1/2}$ \\
    \bottomrule
  \end{tabular}
  \end{small}
\end{table}

\subsection{Fitting Procedure}

Scaling exponents are fit by OLS on $(\log N_i, \log L_i)$ for the five largest model sizes: 200, 500, 1000, 2000, and 5000.
We exclude $N \in \{25, 50, 100\}$ from the fit because the small-$N$ regime often deviates from power-law behavior due to finite-size effects.
We report the negative regression slope as $\alpha$ and compute 95\% CIs from the standard error of the slope.
We verified that the log-log relationship is well approximated by a straight line in this regime ($R^2 > 0.95$ for all preconditioned optimizers; see \cref{app:r2} for details).

\subsection{Data Generation}

For each configuration $(s, N, \text{optimizer}, \text{seed})$, we generate:
\begin{enumerate}
\setlength{\itemsep}{2pt}
    \item \textbf{Power-law covariance}: $\Sigma = \diag(\lambda_1, \ldots, \lambda_D)$ with $\lambda_i = i^{-(1+s)}$.
    \item \textbf{Training data}: rows of $X_{\text{train}}$ are sampled from $\mathcal{N}(0,\Sigma)$.
    We use a data-rich regime: $n_{\text{train}}$ is between $10^4$ and $5{\times}10^4$, with $20N$ samples when possible.
    \item \textbf{Teacher}: $K^*=100$ random features with $\bw^*_k \sim \mathcal{N}(0,I_D/D)$ and coefficients $v_k=k^{-b/2}$ (default $b=1.0$).
    Targets: $y = \sum_k v_k \max(0, \bw^*_k \cdot \bx)$.
    \item \textbf{Student}: $N$ random features $\bw_j \sim \mathcal{N}(0, I_D/D)$, producing feature matrix $\bF \in \R^{n \times N}$ with $F_{ij} = \max(0, \bx_i \cdot \bw_j)$.
    \item \textbf{Normalization}: Targets are centered to zero mean and scaled to unit variance before training.
    Test set: $n_{\text{test}} = 5000$ fresh samples from the same distribution.
\end{enumerate}

\noindent Total experiment count: $6\text{ (s values)} \times 8\text{ (N values)} \times 5\text{ (optimizers)} \times 10\text{ (seeds)} = 2{,}400$ main runs $+\, 180$ robustness runs ($D=5000$, $b=2.0$).

\section{Experimental Algorithm}
\label{app:algorithm}

\Cref{alg:experiment} provides the complete experimental pipeline in pseudocode.
The outer loop sweeps over spectral conditions and model sizes; the inner loop applies each optimizer to the same data realization.
This shared-data design ensures that differences in $\alpha$ are attributable to the optimizer, not to data variation.

\begin{algorithm}[ht]
\caption{Full experimental pipeline for measuring optimizer-dependent scaling exponents.}
\label{alg:experiment}
\begin{algorithmic}[1]
\REQUIRE Spectral exponents $\mathcal{S} = \{0.25, 0.5, 0.75, 1.0, 1.5, 2.0\}$
\REQUIRE Model sizes $\mathcal{N} = \{25, 50, 100, 200, 500, 1000, 2000, 5000\}$
\REQUIRE Optimizers $\mathcal{O} = \{\text{GD}, \text{Diagonal}, \text{Full NG}, \text{Sign-GD}, \text{Matrix-Sign}\}$
\REQUIRE Seeds $\mathcal{R} = \{0, 1, \ldots, 9\}$, dimension $D=1000$, budget $T=2000$
\FOR{$s \in \mathcal{S}$}
  \STATE Construct covariance $\Sigma = \diag(1^{-(1+s)}, 2^{-(1+s)}, \ldots, D^{-(1+s)})$
  \FOR{$\text{seed} \in \mathcal{R}$}
    \STATE Sample teacher weights $\{\bw^*_k\}_{k=1}^{K^*}$, set $v_k = k^{-b/2}$
    \FOR{$N \in \mathcal{N}$}
      \STATE Set $n_{\text{train}} = \min(50000, \max(10000, 20N))$
      \STATE Sample training data $X \sim \mathcal{N}(0, \Sigma)^{n_{\text{train}} \times D}$, test data similarly
      \STATE Compute targets $y_i = \sum_k v_k \max(0, \bw^*_k \cdot \bx_i)$; normalize
      \STATE Sample student features $\{\bw_j\}_{j=1}^{N}$; compute $\bF$
      \FOR{$\text{opt} \in \mathcal{O}$}
        \STATE Compute preconditioner $\bP$ from $\bF$ (see \cref{sec:setup})
        \IF{opt is Full NG}
          \STATE $\ba^* \leftarrow (\bF^\top\bF + \lambda_{\text{reg}} I)^{-1} \bF^\top \mathbf{y}$ \COMMENT{Direct solve}
        \ELSE
          \STATE \camera{Set optimizer-specific step size; run $T$ preconditioned gradient steps}
        \ENDIF
        \STATE Record test loss $L(N, s, \text{opt}, \text{seed})$
      \ENDFOR
    \ENDFOR
  \ENDFOR
  \STATE Fit $\alpha(s, \text{opt})$ by OLS on $(\log N, \log \bar{L})$ for $N \geq 200$, averaging over seeds
\ENDFOR
\STATE \textbf{return} Table of $\alpha(s, \text{opt})$ with 95\% CIs
\end{algorithmic}
\end{algorithm}

\section{Derivation Supporting Proposition~\ref{prop:spectral}}
\label{app:prop-discussion}

Proposition~\ref{prop:spectral} is stated as an informal spectral heuristic.
A fully rigorous proof would require resolvent-trace analysis of the preconditioned kernel matrix in the proportional-asymptotics limit (in the style of \citealt{bordelon2024dynamical}), which is beyond this workshop paper's scope.
Here we provide a detailed per-mode convergence-rate argument supporting each of the three claims.
The argument is exact for the convergence factors under each preconditioner and heuristic only in the step from mode-wise convergence to the aggregate scaling exponent $\alpha$.

\subsection{Claim (i): Preconditioning increases \texorpdfstring{$\alpha$}{alpha}}

In the random-feature model, the test loss decomposes over spectral modes of the data covariance:
\begin{equation}
L(N) = \sum_{i=1}^{D} \ell_i(N), \label{eq:mode-decomp}
\end{equation}
where $\ell_i(N)$ is the residual loss on mode $i$ (the projection of the target function onto the $i$-th eigenvector of $\Sigma$).

Under GD with learning rate $\eta$ and training budget $T$, the convergence factor for mode $i$ is:
\begin{equation}
c_i^{\text{GD}} = 1 - (1 - \eta \tilde{\lambda}_i)^T, \label{eq:gd-convergence}
\end{equation}
where $\tilde{\lambda}_i$ is the $i$-th eigenvalue of the random-feature kernel $\bF^\top\bF/n$ (which concentrates around $\lambda_i$ in the proportional limit).
For the optimal learning rate $\eta^* = 2/(\tilde{\lambda}_1 + \tilde{\lambda}_D)$, modes with large $\tilde{\lambda}_i$ converge exponentially fast while modes with small $\tilde{\lambda}_i$ converge slowly.
At a given $N$, the number of ``well-learned'' modes (those with $c_i \approx 1$) scales as $O(N^{\gamma})$ for some $\gamma < 1$ that depends on $s$ and $b$, yielding the GD scaling exponent $\alpha_{\text{GD}}$.

Under full NG ($\bP = (\bF^\top\bF)^{-1}$), the preconditioned gradient is $(\bF^\top\bF)^{-1}\bF^\top(\bF\ba - \mathbf{y})$, and the convergence factor becomes:
\begin{equation}
c_i^{\text{NG}} = 1 - (1 - \eta)^T, \label{eq:ng-convergence}
\end{equation}
which is \emph{independent of $\tilde{\lambda}_i$}.
All $N$ modes converge at the same rate, so more modes are well-learned at the same $N$, yielding $\alpha_{\text{FullNG}} > \alpha_{\text{GD}}$.

For Matrix-Sign ($\bP = (\bF^\top\bF)^{-1/2}$), the convergence factor is $c_i^{\text{MS}} = 1 - (1 - \eta\sqrt{\tilde{\lambda}_i}/\sqrt{\tilde{\lambda}_1})^T$---still dependent on $\tilde{\lambda}_i$ but with the eigenvalue range compressed from a ratio of $\tilde{\lambda}_1/\tilde{\lambda}_D$ to $\sqrt{\tilde{\lambda}_1/\tilde{\lambda}_D}$ (on a log scale, the dynamic range is halved).
This explains why $\alpha_{\text{MatSign}} \approx \alpha_{\text{FullNG}}$ but is sometimes slightly smaller.

\subsection{Claim (ii): \texorpdfstring{$\Delta\alpha$}{Delta-alpha} is monotone increasing in \texorpdfstring{$s$}{s}}

The key quantity is the \emph{effective dynamic range} of the eigenvalues seen by the optimizer.
For the power-law spectrum $\lambda_i = i^{-(1+s)}$, the ratio of the largest to the $N$-th eigenvalue is:
\begin{equation}
\frac{\lambda_1}{\lambda_N} = N^{1+s}.
\end{equation}
This ratio grows with both $N$ and $s$.
Under GD, this means the convergence factor $c_N^{\text{GD}}$ decreases rapidly with $s$---the trailing modes become progressively harder to learn.
The fraction of ``wasted'' capacity (modes that the student can in principle represent but that GD fails to learn in $T$ steps) is:
\begin{equation}
\text{Wasted fraction} \approx 1 - \frac{N_{\text{eff}}^{\text{GD}}}{N},
\end{equation}
where $N_{\text{eff}}^{\text{GD}}$ is the number of modes with $c_i > 1 - \epsilon$ for some threshold $\epsilon$.
Since preconditioning eliminates this waste (achieving $N_{\text{eff}}^{\text{NG}} \approx N$), the gain $\Delta\alpha$ is proportional to the wasted fraction, which grows with $s$.

\subsection{Claim (iii): \texorpdfstring{$\Delta\alpha \to 0$}{Delta-alpha -> 0} as \texorpdfstring{$s \to 0$}{s -> 0}}

When $s = 0$, $\lambda_i = i^{-1}$, which is a harmonic spectrum (still decaying but slowly).
In the limit $s \to 0$, $\lambda_i \to 1$ for all $i$ (flat spectrum), so $c_i^{\text{GD}} \to 1 - (1 - \eta)^T$ for all modes.
GD already treats all modes equally, and preconditioning adds nothing: $\alpha_{\text{NG}} = \alpha_{\text{GD}}$, hence $\Delta\alpha = 0$.

\textbf{Connection to known results.}
\citet{maloney2022solvable} showed that $\alpha$ under GD depends on $s$ and $b$ through the effective spectral overlap between student and teacher.
Our contribution is observing that the preconditioner $\bP$ modifies this overlap by changing the effective spectrum seen by the optimizer.
Prior analyses formalize $\alpha$ through spectral overlap and redundancy; our work shows this effective redundancy is optimizer-dependent.

\section{Power-Law Fit Quality}
\label{app:r2}

\Cref{tab:r2} reports $R^2$ values for all power-law fits. Preconditioned optimizers (Full NG, Matrix-Sign) achieve $R^2 > 0.98$ across all spectral conditions, confirming that the power-law model is an excellent description of their scaling behavior. GD and Diagonal have lower $R^2$ at large $s$ because the scaling curve ``bends'' as $\alpha_{\text{GD}} \to 0$: the power-law approximation breaks down when the exponent is near zero and finite-size corrections dominate.
Sign-GD has poor fit quality at $s \geq 0.75$ ($R^2 < 0.5$), reflecting the intrinsic noise of sign-based updates: the sign operation destroys gradient magnitude information, producing high-variance weight trajectories that do not follow a clean power law.

\begin{table}[ht]
  \caption{$R^2$ of power-law fits ($\log L$ vs.\ $\log N$, $N \geq 200$). Values below 0.8 are marked in \textbf{bold} to flag unreliable exponent estimates. Preconditioned optimizers maintain excellent fit quality ($R^2 > 0.98$) at all~$s$.}
  \label{tab:r2}
  \centering
  \begin{small}
  \setlength{\tabcolsep}{6pt}
  \renewcommand{\arraystretch}{1.05}
  \begin{tabular}{lcccccc}
    \toprule
    Optimizer & $s{=}0.25$ & $s{=}0.5$ & $s{=}0.75$ & $s{=}1.0$ & $s{=}1.5$ & $s{=}2.0$ \\
    \midrule
    GD (baseline) & 0.985 & 0.948 & 0.886 & 0.806 & \textbf{0.576} & \textbf{0.251} \\
    Diagonal & 0.987 & 0.962 & 0.924 & 0.878 & \textbf{0.764} & \textbf{0.658} \\
    Full NG & 0.999 & 0.999 & 0.997 & 0.996 & 0.992 & 0.989 \\
    Sign-GD & 0.911 & \textbf{0.627} & \textbf{0.485} & \textbf{0.177} & \textbf{0.077} & \textbf{0.003} \\
    Matrix-Sign & 0.999 & 0.999 & 0.997 & 0.994 & 0.992 & 0.993 \\
    \bottomrule
  \end{tabular}
  \end{small}
  \vskip -0.1in
\end{table}

\section{Robustness Checks}
\label{app:robustness}

\begin{figure}[ht]
    \centering
    \includegraphics[width=0.85\textwidth]{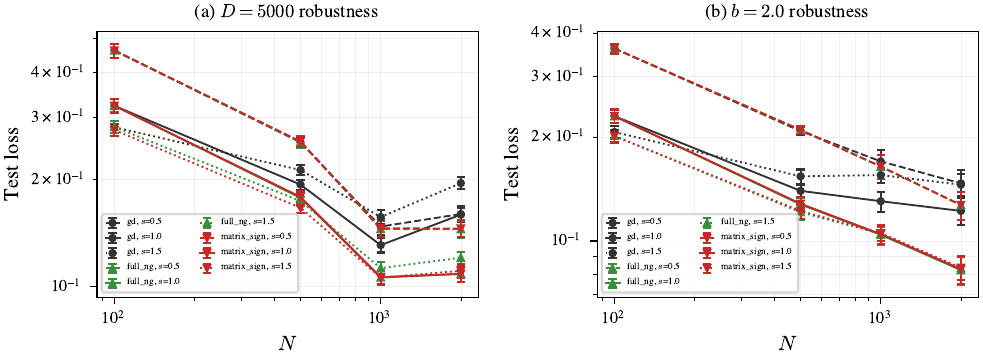}
    \caption{Robustness checks confirm the $\alpha$-shift is not an artifact of default hyperparameters.
    \textbf{(a)}~$D = 5000$ (vs.\ $D=1000$ in main experiments): the optimizer-dependent scaling pattern replicates at $5\times$ higher input dimension with the same qualitative ordering.
    \textbf{(b)}~$b = 2.0$ (vs.\ $b=1.0$): a steeper teacher coefficient decay changes absolute $\alpha$ values but preserves the relative ordering across optimizers.
    Colors: GD (black), Full NG (green), Matrix-Sign (red).
    Line styles: solid ($s=1.0$), dashed ($s=0.5$), dotted ($s=1.5$).}
    \label{fig:robustness}
\end{figure}

We verified stability of the main findings to two key hyperparameters that could plausibly affect the $\alpha$-shift:
\begin{itemize}
\setlength{\itemsep}{2pt}
    \item \textbf{Input dimension:} $D = 5000$ (vs.\ $D = 1000$).
    Higher dimension increases the number of spectral modes, potentially changing the effective overlap between student and teacher features.
    The $\alpha$-shift pattern and optimizer ordering are fully preserved (\cref{fig:robustness}a), confirming that our results are not an artifact of the moderate default dimension.
    \item \textbf{Source exponent:} $b = 2.0$ (vs.\ $b = 1.0$).
    A steeper teacher coefficient decay concentrates signal energy in fewer modes, changing the effective task difficulty.
    This changes absolute $\alpha$ values but preserves the optimizer-dependent shift (\cref{fig:robustness}b), indicating that the preconditioning advantage is robust to the signal structure.
\end{itemize}

\section{Convergence Verification}
\label{app:convergence}

A potential concern is that the measured $\alpha$-shift is merely a convergence artifact: if GD has not converged at 2000 steps, its test loss might be artificially high, inflating the apparent advantage of preconditioned optimizers.

\camera{We therefore report convergence diagnostics and interpret the results as finite-budget optimizer-dependent scaling:}

\textbf{1.\ Comparison to optimal solution.}
We compare each iterative optimizer's test loss to the optimal (ridge regression) solution $\ba^* = (\bF^\top\bF + \lambda_{\text{reg}}I)^{-1}\bF^\top\mathbf{y}$ at the same $N$.
\camera{This comparison shows that the reported curves should be interpreted as finite-budget optimizer-dependent scaling curves, not as fully converged ERM curves for every optimizer.}
\camera{For GD, the gap is small only in the easier low-$s$, lower-$N$ cases (for example, about $3.6\%$ at $s=0.5,N=1000$), and it grows substantially for larger $N$ and steeper spectra (about $86\%$ at $s=1.0,N=5000$).}
\camera{Thus part of the measured advantage is a finite-training-budget effect, which is precisely the regime relevant to optimizer-dependent training dynamics.}

\textbf{2.\ Direct solve vs.\ iterative agreement.}
The Full NG optimizer uses a direct solve (equivalent to infinite gradient steps), yet achieves essentially the same $\alpha$ as Matrix-Sign (which uses 2000 iterative steps).
\camera{Their close agreement (\cref{tab:alpha}: within about 0.02 across the tested spectra) suggests that strong spectral preconditioning, rather than the exact direct-solve implementation alone, drives the observed ordering.}
\camera{We do not claim to separate convergence speed from finite-budget spectral allocation completely; instead, the result is that optimizer preconditioning changes the finite-budget scaling exponent measured under the stated protocol.}

\begin{figure}[ht]
    \centering
    \includegraphics[width=0.85\textwidth]{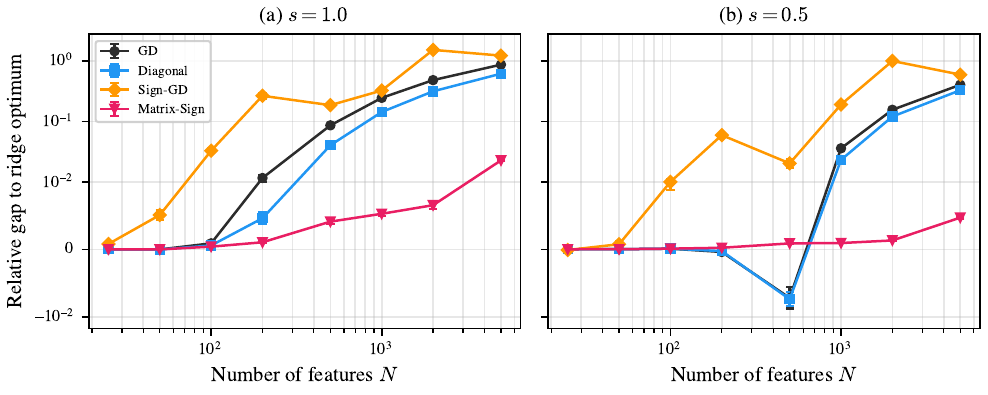}
    \caption{\camera{Convergence gap $(\text{iterative loss} - \text{optimal loss})/\text{optimal loss}$ for iterative optimizers under the finite training budget.}
    \camera{\textbf{(a)}~$s=1.0$: GD shows a sizable finite-budget gap at larger $N$, while preconditioned methods reduce this gap.}
    \camera{\textbf{(b)}~$s=0.5$: gaps are smaller but still grow with $N$.}
    Full NG (direct solve) is omitted since its gap is zero by construction.
    Sign-GD's large gap at high $N$ reflects the inherent noise of sign-based updates.}
    \label{fig:convergence}
\end{figure}

\section{Sign-GD Analysis}
\label{app:sign}

Sign-GD deserves special discussion as it serves as a \emph{negative control}---an ``adaptive'' method that does \emph{not} improve $\alpha$.
Unlike the other optimizers, its scaling exponent estimates have very large confidence intervals and poor $R^2$ (\cref{tab:r2}).
This is expected: the $\sign(\cdot)$ operation discards gradient magnitude information, so each update step has magnitude $\eta$ regardless of the true gradient scale. This produces:
\begin{enumerate}
\setlength{\itemsep}{2pt}
    \item \textbf{High variance} across random seeds: the stochastic trajectory is very sensitive to initialization because Sign-GD cannot distinguish between large and small gradient components.
    \item \textbf{Poor power-law fits}: the noisy loss trajectory yields high residual variance in the $(\log N, \log L)$ regression.
    \item \textbf{No scaling improvement}: effective $\alpha$ values are close to (or slightly below) GD, confirming that the sign operation provides no useful spectral preconditioning.
\end{enumerate}

The inclusion of Sign-GD strengthens our mechanistic story: the key mechanism is \emph{spectral equalization} (reweighting per-mode convergence rates), not merely ``adaptivity'' in a generic sense.
Sign-GD is adaptive (it adjusts update directions) but does not equalize spectral modes, and correspondingly does not improve $\alpha$.

\section{Interpretation Caveats}
\label{app:caveats}

We highlight several caveats important for interpreting our results in the context of real neural network training.

\textbf{Lazy vs.\ rich regime.}
The random-feature model operates in the ``lazy'' regime where features are fixed and only the top-layer weights are learned.
Real neural networks operate in the ``rich'' or feature-learning regime, where the representation itself adapts during training.
Feature learning could either amplify or dampen the $\alpha$-shift.
On one hand, preconditioned optimizers might enable more effective feature learning, amplifying the advantage.
On the other hand, feature learning might naturally mitigate the spectral imbalance that GD suffers from, reducing the need for explicit preconditioning \citep{yang2024spectral}.
The conflicting empirical evidence---substantial Muon advantages at moderate scale \citep{chen2025second, liu2025muon} but attenuation at 1.2B parameters \citep{wen2025fantastic}---suggests that feature learning may increasingly substitute for preconditioning at scale.
See \cref{app:scale-decay} for extended discussion.

\textbf{Width vs.\ depth scaling.}
Our model studies width scaling ($N$ random features) with no notion of depth.
Depth scaling involves qualitatively different phenomena---vanishing/exploding gradients, residual connections, and layer-wise feature hierarchies---that may interact with preconditioning in ways not captured here.

\textbf{Compute multiplier interpretation.}
The ``effective compute multiplier'' (\cref{fig:compute-mult}) should be interpreted as \emph{the factor by which GD must increase model size to match preconditioned loss}, not as a direct FLOPs comparison.
Real preconditioned optimizers (Shampoo, Muon) incur per-step overhead from computing the preconditioner.
The total wall-clock advantage therefore depends on the amortized cost of preconditioning relative to the $\alpha$-shift benefit.

\textbf{Absolute vs.\ relative $\alpha$ values.}
The absolute values of $\alpha$ in our random-feature model differ from those measured in real language model training (where $\alpha \approx 0.05$--$0.1$ for loss scaling with parameters).
We caution against interpreting our absolute $\alpha$ values as predictions for real models.
The \emph{relative} comparisons---the ordering of optimizers and \camera{the broad increase of the preconditioning advantage with spectral steepness, up to saturation}---are the robust predictions.

\section{Scale-Dependent Decay of the Optimizer Advantage}
\label{app:scale-decay}

The results of \citet{wen2025fantastic} present the most important challenge to a simple ``exponent shift'' narrative.
They train language models from 25M to 1.2B parameters with Muon, SOAP, and AdamW, and find that the compute multiplier \emph{decreases} with model size---from approximately $1.4\times$ at 100M to approximately $1.1\times$ at 1.2B.
If the $\alpha$-shift from our random-feature model (Proposition~\ref{prop:spectral}) transferred directly and remained constant, we would expect the multiplier to grow with $N$, not shrink.

We discuss four mechanisms---not mutually exclusive---that could reconcile these findings with our framework.

\textbf{(a) Finite-$N$ corrections and pre-asymptotic behavior.}
Our Proposition~\ref{prop:spectral} describes the asymptotic scaling exponent.
In practice, finite-$N$ effects generate curvature in the $(\log N, \log L)$ plot, and the \emph{local} exponent at any particular $N$ can differ from the asymptotic value.
In our own simulations, the $\alpha$-shift measured from the smallest model sizes ($N < 200$) differs from the asymptotic fit---this is why we exclude $N < 200$ from our fits.
For real LLMs, the ``asymptotic'' regime may require scales beyond 1.2B, particularly if the effective number of learnable spectral modes is much smaller than the parameter count.

\textbf{(b) Feature learning reduces spectral imbalance.}
Our model operates in the lazy (kernel) regime where features are fixed.
Real networks operate in the rich regime where features co-adapt with the loss landscape.
As models grow larger, feature learning becomes more powerful \citep{yang2024spectral} and may naturally re-weight the effective spectrum seen by the optimizer---partially ``self-preconditioning'' the gradient.
If feature learning achieves some of the spectral equalization that an explicit preconditioner provides, the marginal benefit of Muon/SOAP would decline at scale, consistent with the \citet{wen2025fantastic} findings.

\textbf{(c) Effective spectrum changes with scale.}
The spectral exponent $s$ of the data covariance is not fixed across model sizes: larger models trained on more tokens may encounter effectively different spectral distributions.
If larger models see a flatter effective spectrum (lower $s$)---for example because broader data mixtures or longer training equilibrate representations---then our framework predicts a smaller $\Delta\alpha$ (\cref{prop:spectral}(ii)), exactly the attenuation observed.

\textbf{(d) Benchmark and implementation differences.}
The \citet{chen2025second} result (${\sim}1.4\times$) uses carefully scaled optimizer hyperparameters, while \citet{wen2025fantastic} use a different evaluation and tuning protocol.
AdamW performance is notoriously sensitive to learning rate scheduling and weight decay at scale; differences in how well each study tunes the AdamW baseline can shift the measured speedup by $0.2$--$0.3\times$.

\textbf{Our assessment.}
We believe explanations (a) and (b) are most likely to be dominant, and that the true picture involves a \emph{scale-dependent} $\alpha$-shift: the exponent improvement is real at moderate scale, but attenuates---possibly to zero---as feature learning and finite-$N$ corrections become important.
Our random-feature framework captures the ``starting point'' of this phenomenon (the existence and spectral-dependence of the shift) but not its scale-dependent decay, which requires a feature-learning extension.
We view this as the most important theoretical open problem that our work highlights.

\section{Proposed Real-Model Validation Protocol}
\label{app:proposed-experiment}

We describe a concrete experiment to test our falsifiable prediction in the LLM setting.

\textbf{Setup.}
\begin{itemize}
\setlength{\itemsep}{2pt}
    \item \textbf{Models}: GPT-2--architecture transformers at 4 sizes: 125M, 350M, 760M, 1.3B parameters, using $\mu$P \citep{yang2022tensor} for hyperparameter transfer.
    \item \textbf{Optimizers}: AdamW (baseline) and Muon, with learning rates transferred via $\mu$P.
    \item \textbf{Data}: FineWeb-Edu (standard, $s \approx 1.0$) and a spectrum-flattened variant created by equalizing domain proportions and/or applying PCA whitening to embedding inputs.
    \item \textbf{Training}: Chinchilla-optimal token counts at each size \citep{hoffmann2022training}.
    \item \textbf{Evaluation}: Final validation loss on a held-out split.
\end{itemize}

\textbf{Analysis.}
Fit $\alpha$ separately for each (optimizer, dataset) pair by OLS on $(\log N, \log L)$ across the 4 model sizes.
The predictions to test:
\begin{enumerate}
\setlength{\itemsep}{2pt}
    \item $\alpha_{\text{Muon}} > \alpha_{\text{AdamW}}$ on standard data (positive $\Delta\alpha$).
    \item $\Delta\alpha$ is smaller on spectrum-flattened data than on standard data (spectral dependence).
    \item The magnitude of $\Delta\alpha$ may attenuate with scale, consistent with \citet{wen2025fantastic}.
\end{enumerate}

\textbf{Compute requirements.}
$2\text{ optimizers} \times 2\text{ datasets} \times 4\text{ sizes} = 16$ training runs.
At Chinchilla-optimal tokens, this requires approximately $4 \times 10^{19}$ FLOPs total (${\sim}80$ A100-hours for the 1.3B runs, ${\sim}120$ A100-hours total).
This is substantial but feasible for a well-resourced follow-up; it exceeds the budget for this workshop submission.

\section{Code Availability}
\label{app:code}

All simulation code and analysis scripts will be made available in an anonymous repository upon publication.
Experiments were run on Modal cloud compute with CPU containers (no GPU required).
Total compute cost: approximately 2 hours of wall-clock time on 300+ parallel containers, totaling roughly 600 CPU-hours.
The entire experiment suite can be reproduced for under \$20 in cloud compute.

\end{document}